\documentclass[letterpaper, 10 pt, conference]{ieeeconf}  %

\IEEEoverridecommandlockouts                              %

\usepackage[space]{cite}
\usepackage{amsmath,amssymb,amsfonts}
\usepackage{algorithmic}
\usepackage{graphicx}
\usepackage{textcomp}
\usepackage{xcolor}
\usepackage{booktabs}
\usepackage{hyperref}
\usepackage{multirow}

\usepackage[font=footnotesize]{caption}
\usepackage{cleveref} %

\title{\LARGE \bf
Safe, Task-Consistent Manipulation with \\ Operational Space Control Barrier Functions
}

\author{Daniel Morton and Marco Pavone%
\thanks{Daniel Morton was supported by a NASA Space Technology Graduate Research Opportunity}%
\thanks{Daniel Morton and Marco Pavone are with the Departments of Mechanical Engineering and Aeronautics \& Astronautics, Stanford University, Stanford, CA 94305.
        {\tt\small \{dmorton, pavone\}@stanford.edu}}%
}

\begin{document}
\makeatletter

\makeatother
\maketitle
\thispagestyle{empty}
\pagestyle{empty}

\begin{figure*}
    \centering
    \smallskip
    \includegraphics[width=0.85\textwidth]{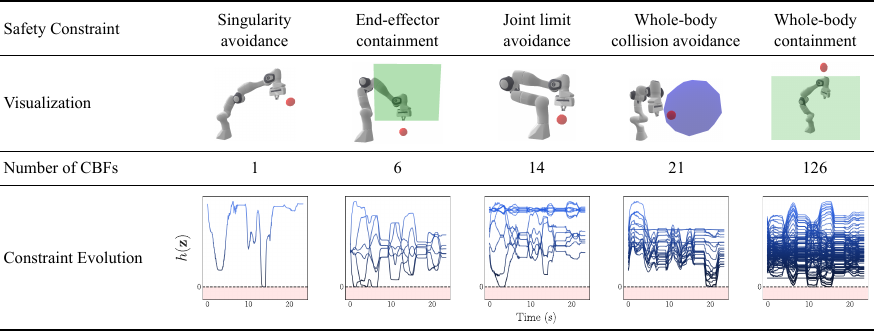}
    \caption{\textbf{Safe and performant highly-constrained manipulation}. Our OSCBF controller maintains safety for hundreds of constraints enforced concurrently, using the full second-order robot dynamics and torque control, and operates at real-time control rates (over 1000 Hz). Shown above is the simultaneous constraint evolution for all CBFs during one adversarial trajectory from teleoperation (end-effector target shown in red). Across 168 safety conditions, the robot remains safe (\(h(\mathbf{z}) >0\)) without over-conservative behavior near the boundary of safety (\(h(\mathbf{z}) =0\)).}
    \label{fig:combined}
    \vspace{-2mm}
\end{figure*}

\begin{abstract}

Safe real-time control of robotic manipulators in unstructured environments requires handling numerous safety constraints without compromising task performance. Traditional approaches, such as artificial potential fields (APFs), suffer from local minima, oscillations, and limited scalability, while model predictive control (MPC) can be computationally expensive. Control barrier functions (CBFs) offer a promising alternative due to their high level of robustness and low computational cost, but these safety filters must be carefully designed to avoid significant reductions in the overall performance of the manipulator. In this work, we introduce an Operational Space Control Barrier Function (OSCBF) framework that integrates safety constraints while preserving \textit{task-consistent} behavior. Our approach scales to hundreds of simultaneous constraints while retaining real-time control rates, ensuring collision avoidance, singularity prevention, and workspace containment even in highly cluttered settings or during dynamic motions. By explicitly accounting for the task hierarchy in the CBF objective, we prevent degraded performance across both joint-space and operational-space tasks, when at the limit of safety. We validate performance in both simulation and hardware, and release our open-source high-performance code and media on our project webpage, {\small\url{https://stanfordasl.github.io/oscbf/}}
\end{abstract}

\section{Introduction}

With autonomous robotic manipulators increasingly operating in unstructured environments, real-time safe control is critical -- not only for collision avoidance but also for a broad range of other safety constraints that are common to manipulators, such as singularity avoidance. An ideal control framework should enforce these safety constraints while minimally modifying the robot's desired behavior, even in dynamic scenarios or when multiple constraints must be satisfied simultaneously.

Enforcing safety while retaining performance is particularly relevant for recent learning-based controllers, which may be quite versatile, but do not provide guarantees on safety \cite{chi2024diffusionpolicy}. Often, these learning-based policies operate in operational space, via end-effector motion commands. Similarly, when collecting data for imitation learning, teleoperated demonstrations often only command a motion of the end-effector \cite{khazatsky2024droid, openx}. In both settings (training and deployment), the manipulator must rely on a lower-level operational space controller (OSC) \cite{KhatibOSC} to map the end-effector motions back to the whole-body control of the robot. This introduces potential safety concerns at multiple levels: the operational space command itself may be unsafe, or the mapping back to the whole-body motion may be unsafe. To maintain safety, the low-level controller must therefore consider multiple constraints, in both the operational and joint spaces.

Artificial potential fields (APFs) \cite{KhatibObstacles} have long served as a simple way of encoding safety in operational space control via repulsive forces around obstacles. However, APFs suffer from a few key problems: they can lead to oscillations when close to obstacles or moving quickly, they can influence the dynamics even far from an unsafe region, and they can require extensive tuning to avoid creating local minima \cite{koren1991potential, SingletaryCBFvsAPF}. APFs can be designed to accommodate multiple task objectives and safety constraints \cite{KhatibConstraintConsistentControl} but this requires an explicit hierarchy and null space for each constraint, which does not scale to highly-constrained settings.

Model predictive control (MPC), alternatively, handles safety via constrained receding-horizon optimal control.
However, MPC can face computational challenges, particularly if the system is subject to nonconvex constraints and nonlinear dynamics. For MPC to run at real-time rates on a manipulator, either safety conditions are not considered \cite{HighFreqNMPCManip}, or assumptions must be made about the reference trajectory \cite{MPCManipulatorSingularityHierarchy}.

Another approach, control barrier functions (CBFs) are a minimally-invasive means of enforcing safety for nonlinear systems via optimization-based controllers, and improve upon APFs with less computational demands than MPC \cite{SingletaryCBFvsAPF}. Due to their robustness and real-time capabilities, CBFs have been successfully demonstrated on a wide range of systems,
 including manipulators \cite{SingletaryFoodPrep, MurtazaTorqueControl, RauscherConstrainedRobotControl, SingularityCBF, OnlineCBFConstructionManipulation, CBFGrasping, OnlineActiveManipulatorSafety}. 

Previous work on CBFs for manipulators often assume a reduced-order model with direct control over the joint velocities, and only extend safety conditions to the full-order dynamics with the assumption that a low-level controller can track a velocity command reasonably well \cite{SingletaryFoodPrep, OnlineCBFConstructionManipulation}. However, this condition does not hold under dynamic maneuvers, and cannot be used with compliant control. Often, these safety criteria are analyzed only over joint-space motions, neglecting the impact on the operational-space dynamics which are critical for teleoperation and learning-based control \cite{OnlineActiveManipulatorSafety}. Works that analyze the operational space \cite{MurtazaTorqueControl, RauscherConstrainedRobotControl} often neglect how to best balance hierarchical task performance and safety in the joint space and operational space, or handling multiple safety constraints. %

\subsection{Statement of Contributions}

We present a framework for integrating Control Barrier Functions (CBFs) into Operational Space Contol (OSC) which (1) enforces safety while respecting the hierarchical task definition of OSC across the operational and joint spaces, (2) applies to both kinematically- and dynamically- controlled manipulators, and (3) scales to large numbers of safety conditions while maintaining real-time control rates, even when performing dynamic motions through highly-occluded environments. Additionally, we release our software, CBFpy \cite{Morton_CBFpy_2024}, which provides an easy-to-use and high-performance interface for CBF controller design. The code, along with additional media, is available at {\small\url{https://stanfordasl.github.io/oscbf/}}

\subsection{Paper Organization}

In Section \ref{sec:CBFs}, we review the construction of control barrier functions for constraints of relative degree 1 and 2, and discuss practical application of CBFs for highly-constrained systems. In Section \ref{sec:manip_kin_dyn}, we review manipulator kinematics, dynamics, and tasks. In Section \ref{sec:oscbf}, we detail the necessity of \textit{task-consistent} safety filters, then we construct our Operational Space Control Barrier Function (OSCBF) controller, for both velocity- and torque- controlled robots. In Section \ref{sec:results}, we present analysis of the OSCBF controller's performance and computational efficiency for five common types of safety constraints, then, in simulation and on hardware, we evaluate the controller in highly constrained environments and during dynamic motions.

\section{\label{sec:CBFs}Control Barrier Functions}

Consider a continuous-time dynamical system in control-affine form:
\begin{equation}
    \dot{\mathbf{z}} = f(\mathbf{z}) + g(\mathbf{z})\mathbf{u} 
\end{equation}
with state \(\mathbf{z} \in \mathcal{Z} \subseteq \mathbb{R}^{n}\), input \(\mathbf{u} \in \mathcal{U} \subseteq \mathbb{R}^m\), and locally Lipschitz continuous dynamics functions \(f : \mathbb{R}^n \rightarrow \mathbb{R}^n\) and \(g : \mathbb{R}^n \rightarrow \mathbb{R}^{n \times m}\).

Safety can be posed through the lens of set invariance. For a safe subset of the state space, \(\mathcal{C} \subset \mathcal{Z}\), if we can define a control barrier function \(h(\mathbf{z}) : \mathbb{R}^n \rightarrow \mathbb{R}\) where \(\mathcal{C}\) is the zero-superlevel set of \(h\), then a controller satisfying
\begin{equation}
    \dot{h}(\mathbf{z, u}) \geq -\alpha(h(\mathbf{z}))
    \label{eq:cbf_condition}
\end{equation}
for \(\mathbf{u} \in \mathcal{U}\) and extended class \(\mathcal{K}_{\infty}\) function \(\alpha\) will render \(\mathcal{C}\) forward-invariant \cite{CBFTheoryAndApplications}.

This condition, Eq. \ref{eq:cbf_condition}, can be integrated into a quadratic program (QP) convex optimization problem, paired with a min-norm objective to operate as a safety filter on a nominal (unsafe) controller:
\begin{equation}
    \begin{aligned}
        \underset{\mathbf{u}}{\text{minimize}} \quad & \| \mathbf{u} - \mathbf{u}_{\text{nom}} \|_2^2 \\
        \text{subject to} \quad & L_f h(\mathbf{z}) + L_g h(\mathbf{z}) \mathbf{u} \geq -\alpha\left( h(\mathbf{z}) \right)
    \end{aligned}
    \label{eq:cbf}
\end{equation}
where \(L_f\) and \(L_g\) are the Lie derivatives of \(h\) along the dynamics, and \(\dot{h}(\mathbf{z, u}) = L_f h(\mathbf{z}) + L_g h(\mathbf{z}) \mathbf{u}\). 

The \textit{relative degree} of a CBF refers to the number of differentiations along the dynamics required before the control input \(\mathbf{u}\) explicitly appears. CBFs require a relative degree of 1 (RD1), but with mechanical systems, CBFs are often of relative degree 2 (RD2) \cite{CBFGrasping}. This high relative degree implies that \(L_g h(\mathbf{z}) = 0\), and thus, we must differentiate along the dynamics again. The second time-derivative of \(h(\mathbf{z})\) can be expressed as
\begin{equation}
    \ddot{h}(\mathbf{z}, \mathbf{u}) = L_{f}^{2}h(\mathbf{z}) + L_{g}L_{f}h(\mathbf{z})\mathbf{u}
\end{equation}
where \(L_g L_f h(\mathbf{z}) \neq 0\) for a RD2 CBF. 

Given this, we can construct a High-Order CBF (HOCBF) \cite{XiaoHOCBFs} for these RD2 constraints. Let \(h_2(\mathbf{z}) = L_fh(\mathbf{z}) + \alpha(h(\mathbf{z}))\). Then, we modify the constraint in Eq. \ref{eq:cbf} to
\begin{equation}
    L_f h_2(\mathbf{z}) + L_g h_2(\mathbf{z}) \mathbf{u} \geq -\alpha_2\left( h_2(\mathbf{z}) \right)
    \label{eq:rd2_cbf_constraint}
\end{equation}
for an additional class \(\mathcal{K}_\infty\) function \(\alpha_2\)

\begin{table}
    \centering
    \smallskip
    \caption{Safety Conditions: QP Constraint or CBF}
    \label{tab:constraint_or_cbf}
    \begin{tabular}{@{}lll@{}}
    \toprule
    Safety Condition             & Velocity Control & Torque Control \\ \midrule
    Joint position limit         & RD1 CBF          & RD2 CBF        \\
    Joint velocity limit         & Constraint       & RD1 CBF        \\
    Joint torque limit           & ---              & Constraint     \\
    Operational position limit   & RD1 CBF          & RD2 CBF        \\
    Operational velocity limit   & Constraint       & RD1 CBF        \\
    Operational wrench limit     & ---              & Constraint     \\
    Singularity avoidance        & RD1 CBF          & RD2 CBF        \\
    Collision avoidance          & RD1 CBF          & RD2 CBF        \\ \bottomrule
    \end{tabular}
    \vspace{-2mm}
\end{table}

An overview of the relative degree of various safety conditions for manipulators can be found in Table \ref{tab:constraint_or_cbf}.

\textit{Remark}: When enforcing a large number of CBFs, particularly with input constraints, these will sometimes be in conflict, resulting in an infeasible QP. Finding valid, non-overly-conservative CBFs is an ongoing research challenge \cite{BreedenHighRelativeDegInputConstraints}, but in practice, relaxing the QP results in a reasonable solution that enforces (but does not guarantee) safety in most cases.

Given this, we can relax the QP (Eq. \ref{eq:cbf}) by introducing a slack variable, \(\mathbf{t}\) to handle the constraint violation with a large penalty, \(\rho\):
\begin{equation}
    \begin{aligned}
        \underset{\mathbf{u}}{\text{minimize}} \quad & \| \mathbf{u} - \mathbf{u}_{\text{nom}} \|_2^2  + \rho^T \mathbf{t}\\
        \text{subject to} \quad & L_f h(\mathbf{z}) + L_g h(\mathbf{z}) \mathbf{u} \geq -\alpha\left( h(\mathbf{z}) \right) - \mathbf{t} \\ 
        & \mathbf{t} \geq \mathbf{0}
    \end{aligned}
    \label{eq:relaxed_cbf}
\end{equation}

\section{\label{sec:manip_kin_dyn}Manipulator Control}

\subsection{Manipulator Kinematics}

We consider a serial-chain manipulator with a kinematic model of the following form:
\begin{equation}
    \boldsymbol{\nu}(\mathbf{q}, \dot{\mathbf{q}}) = J(\mathbf{q})\dot{\mathbf{q}}
    \label{eq:kinematic_model}
\end{equation}
where the Jacobian \(J\) maps joint velocities \(\dot{\mathbf{q}} \in \mathbb{R}^{n_q}\) to the operational space twist \(\boldsymbol{\nu} = \left[\mathbf{\dot{x}}_p , \boldsymbol{\omega}\right] \in \mathbb{R}^{6}\).

We denote a generalized inverse of the Jacobian \( J\) as \( J^{\#}\), and for a non-redundant manipulator, \(J^{\#} = J^{-1}\). Given this, the relationship between operational space twist and joint velocities can be expressed as:
\begin{equation}
    \dot{\mathbf{q}} = J^{\#}(\mathbf{q})\boldsymbol{\nu} + N(\mathbf{q})\dot{\mathbf{q}}_0
\end{equation}
where \(N(\mathbf{q})\) is the null space projection matrix associated with \(J^{\#}\), 
\begin{equation}
    N(\mathbf{q}) = I - J^{\#}(\mathbf{q})J(\mathbf{q})
\end{equation}
and \(\dot{\mathbf{q}}_0\) is an arbitrary joint velocity vector.

\subsection{Manipulator Dynamics}

We consider a serial-chain manipulator with a dynamics model of the following form:
\begin{equation}
    M(\mathbf{q}) \mathbf{\ddot{q}} + \mathbf{c}(\mathbf{q}, \dot{\mathbf{q}}) + \mathbf{g}(\mathbf{q}) = \boldsymbol{\Gamma}
\end{equation}

where \(M(\mathbf{q})\) is the mass matrix, \(\mathbf{c}(\mathbf{q}, \dot{\mathbf{q}})\) is the vector of centrifugal and Coriolis forces, \(\mathbf{g}(\mathbf{q})\) is the gravity vector, and \(\boldsymbol{\Gamma}\) is the vector of joint torques. 

Likewise, we can describe the dynamic behavior of the manipulator's end-effector in operational space as follows:
\begin{equation}
    \Lambda(\mathbf{q}) \dot{\boldsymbol{\nu}} + \boldsymbol{\mu}(\mathbf{q}, \dot{\mathbf{q}}) + \mathbf{p}(\mathbf{q}) = \boldsymbol{\mathcal{F}}
\end{equation}

where \(\Lambda(\mathbf{q})\) is the operational space mass matrix, \(\boldsymbol{\mu}(\mathbf{q}, \dot{\mathbf{q}})\) is the operational space centrifugal and Coriolis force vector, \(\mathbf{p}(\mathbf{q})\) is the operational space gravity vector, and \(\boldsymbol{\mathcal{F}}\) is the operational space wrench.

The components of the joint-space and operational-space dynamic models are related via the following:
\begin{equation}
    \Lambda(\mathbf{q}) = (J(\mathbf{q}) M^{-1}(\mathbf{q}) J^{T}(\mathbf{q}))^{-1}
\end{equation}
\begin{equation}
    \boldsymbol{\mu}(\mathbf{q}, \dot{\mathbf{q}}) = \bar{J}^{T}(\mathbf{q}) \mathbf{c}(\mathbf{q}, \dot{\mathbf{q}})) - \Lambda(\mathbf{q})\dot{J}(\mathbf{q}, \dot{\mathbf{q}})\dot{\mathbf{q}} 
\end{equation}
\begin{equation}
    \mathbf{p}(\mathbf{q}) = \bar{J}^{T}(\mathbf{q})\mathbf{g}(\mathbf{q})
\end{equation}
where \(\bar{J}(\mathbf{q})\) is the dynamically-consistent generalized inverse Jacobian for a redundant manipulator, 
\begin{equation}
    \bar{J}(\mathbf{q}) = M^{-1}(\mathbf{q}) J^{T}(\mathbf{q}) \Lambda(\mathbf{q})
\end{equation}

For a non-redundant manipulator, these same equations apply where \(\bar{J}(\mathbf{q}) = J^{-T}(\mathbf{q})\)

The relationship between operational space wrenches and joint space torques can be expressed as:
\begin{equation}
    \boldsymbol{\Gamma} = J^{T}(\mathbf{q}) \boldsymbol{\mathcal{F}} + N^T(\mathbf{q})\boldsymbol{\Gamma}_{0}
\end{equation}
where \(N^T(\mathbf{q})\) is the null space projection matrix associated with \(J^{T}\),
\begin{equation}
    N^T(\mathbf{q}) = I - J^{T}(\mathbf{q}) \bar{J}^{T}(\mathbf{q})
\end{equation}
\(I\) is an identity matrix, and \(\boldsymbol{\Gamma}_{0}\) is an arbitrary joint torque vector which will be projected into the null space of \(J^{T}(\mathbf{q})\). We denote a vector of joint torques in the null space as \(\boldsymbol{\Gamma}_{N}\)

\subsection{Task Hierarchy}

For our experiments, we consider the primary \textit{task} as tracking a desired operational space pose \(\in \mathbb{R}^6\), though other tasks can similarly apply (e.g., position tracking \(\in \mathbb{R}^3\) using the linear Jacobian, \(J_v\)). For a redundant robot, a common secondary task is to maintain a joint posture with high manipulability characteristics, in which case \(N(\mathbf{q})\dot{\mathbf{q}}_0\) (velocity control) or \(N^T(\mathbf{q})\boldsymbol{\Gamma}_0\) (torque control) can be used to approach this posture without affecting the end-effector.

\section{\label{sec:oscbf}Operational Space Control Barrier Functions}

\subsection{Task Consistency}

\begin{figure*}
    \centering
    \smallskip
    \includegraphics[width=0.6\linewidth]{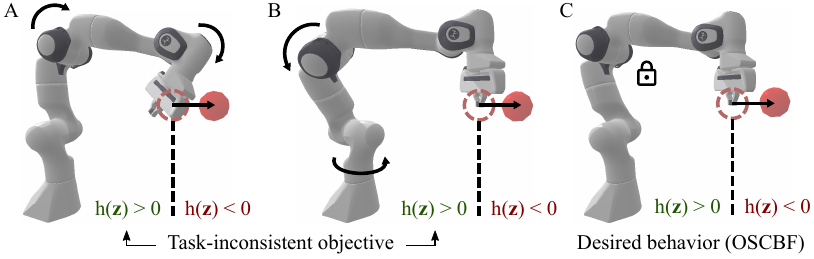}
    \caption{\textbf{Task-consistency: balancing safety and performance}. Consider the behavior of the robot with the tip of the end-effector at the boundary of safety, where the desired goal moves towards the unsafe region. If the CBF objective is not \textit{task-consistent}, the safety filter leads to a decrease in performance, even if safety is maintained. In (A), a joint space metric leads to a decrease in the operational space task performance, and in (B), a purely operational space metric leads to excess motion in the null space. (C) With OSCBF, excess motion is minimized and both safety and task performance is maintained.}
    \label{fig:behavior}
    \vspace{-2mm}
\end{figure*}

When imposing a safety constraint on a robotic manipulator, many trajectories can maintain safety, but not all can additionally maximize task performance. As shown in Fig. \ref{fig:behavior}, while ``task-inconsistent" CBFs can remain safe (i.e., \(h(\mathbf{z}) \geq 0\)), they can also introduce excess motion at the boundary of safety. Examples of ``task-inconsistency" include: (1) optimizing a joint-space metric, while the task is defined in the operational space; (2) the converse: optimizing an operational-space metric, while ignoring a secondary null space joint task; (3) optimizing a torque-based metric, rather than an acceleration-based metric.

In general, \textit{task consistency} implies that rather than applying the CBF safety filter directly to the control input, the filter should minimally modify an output that reflects the task and hierarchy definition. By doing so, this eliminates unnecessary motion when the reference command moves further into the unsafe set. Inconsistent examples (1) and (2) above misrepresent the task hierarchy. (3) optimizes a metric which less directly reflects the position-based tasks than acceleration, due to the additional inertial mapping involved.

\subsection{Kinematic-OSCBF}

For velocity-controlled manipulators, computing the safe control input \(\dot{\mathbf{q}}\) requires three steps: (1) Determine the nominal velocities to reduce error in the task hierarchy; (2) Map the nominal velocities to an (unsafe) joint velocity command; (3) Pass the nominal joint velocities through the OSCBF QP to yield the safe velocity command \(\dot{\mathbf{q}}^*\).

We begin by computing the nominal task velocities, \(\boldsymbol{\nu} \) and \(\dot{\mathbf{q}}_{N}\), where 
\(\boldsymbol{\nu}\) is a proportional operational space twist command \(\in \mathbb{R}^{6}\):
\begin{equation}
    \boldsymbol{\nu} = \boldsymbol{\nu}_{\text{des}} - K_{po}\begin{bmatrix} \mathbf{x}_p - \mathbf{x}_{p,\text{des}}\\  \boldsymbol{\delta\phi} \end{bmatrix}
\end{equation}
and \(\dot{\mathbf{q}}_N\) is a proportional joint velocity command \(\in \mathbb{R}^{n_q}\), projected into the nullspace of the operational space task:
\begin{equation}
    \dot{\mathbf{q}}_N = N(\mathbf{q}) \left(\dot{\mathbf{q}}_{\text{des}} - K_{pj}(\mathbf{q} - \mathbf{q}_{\text{des}})\right)
\end{equation}
Here, \(\boldsymbol{\delta\phi}\) represents the instantaneous angular error vector between \( \mathbf{x}_r\) and \( \mathbf{x}_{r, \text{des}}\). We can easily reconstruct the rotation matrices \(R\) and \(R_{\text{des}}\) from the flattened representation \( \mathbf{x}_r\) and compute this as follows, where \(\mathbf{r}_i\) is the \(i^{th}\) column of \(R\):
\begin{equation}
    \boldsymbol{\delta\phi} = -\frac{1}{2}(\mathbf{r}_1 \times \mathbf{r}_{1,\text{des}} + \mathbf{r}_2 \times \mathbf{r}_{2,\text{des}} + \mathbf{r}_3 \times \mathbf{r}_{3,\text{des}})
\end{equation}

\(K_{pj}\) and \(K_{po}\) are proportional gains for the joint space and operational space, respectively.

Combining these together, the nominal (unsafe) joint velocity command is
\begin{equation}
    \dot{\mathbf{q}}_{\text{nom}} = J^{\#}(\mathbf{q})\boldsymbol{\nu} + \dot{\mathbf{q}}_N
\end{equation}

To construct the CBF, we first define the control-affine joint-space dynamics (\(\mathbf{z} = \mathbf{q} \in \mathbb{R}^n\)), with respect to the joint velocity command (\(\mathbf{u} = \dot{\mathbf{q}} \in \mathbb{R}^m\)). For velocity control, we apply a reduced-order model where we assume direct control of the joint velocities.
\begin{equation}
    \dot{\mathbf{z}} = \mathbf{u}
\end{equation}

With these dynamics, and a CBF \(h\), we can then construct the OSCBF QP:
\begin{equation}
    \begin{aligned}
        \underset{\dot{\mathbf{q}}}{\text{minimize}} \quad &  \| W_j(\dot{\mathbf{q}}_{N} - \dot{\mathbf{q}}_{\text{N,nom}}) \|_2^2 + \| W_o (\boldsymbol{\nu} - \boldsymbol{\nu}_{\text{nom}}) \|_2^2 \\ %
        \text{subject to} \quad & L_f h(\mathbf{z}) + L_g h(\mathbf{z}) \mathbf{u} \geq -\alpha\left( h(\mathbf{z}) \right) \\ %
    \end{aligned}
    \label{eq:velocity_cbf}
\end{equation}

Here, the CBF constraint is paired with a task-consistent objective which minimizes deviations from the nominal joint and operational space velocities. \(W_j\) and \(W_o\) are positive-definite diagonal matrices, which can optionally be used to adjust the objective weighting between joint space and operational space deviations (for instance, to adjust the relative magnitudes between position and orientation components, or between prismatic and revolute joints). 

Equivalently, the objective function of Eq. \ref{eq:velocity_cbf} can be written as
\begin{equation}
    \| W_jN(\mathbf{q})(\dot{\mathbf{q}} - \dot{\mathbf{q}}_{\text{nom}}) \|_2^2 + \| W_o J(\mathbf{q})(\dot{\mathbf{q}} - \dot{\mathbf{q}}_{\text{nom}})\|_2^2
\end{equation}
which we can then use to construct a QP with an objective of form \(\frac{1}{2}\mathbf{x}^T P_{QP} \mathbf{x} + \mathbf{q}_{QP}^T\mathbf{x}\), where
\begin{equation}
    P_{QP} = N^T W_j^T W_j N + J^T W_o^T W_o J
\end{equation}
\begin{equation}
    \mathbf{q}_{QP}^T = -\dot{\mathbf{q}}_{\text{nom}}^T P_{QP}
\end{equation}

While we consider a hierarchy of only \(n_t = 2\) tasks, this objective can also generalize to larger hierarchies, by considering the effect of the safety filter on each task \(i\):
\begin{equation}
    \sum_{i=1}^{n_t} \Vert W_i  (\mathbf{v}_i(\mathbf{q}, \dot{\mathbf{q}}) - \mathbf{v}_{i, \text{nom}}(\mathbf{q}, \dot{\mathbf{q}}_{\text{nom}}))\Vert_2^{2}
    \label{eq:velocity_hierarchy}
\end{equation}
where \(\mathbf{v}_i\) is a velocity, in operational or joint space.

Joint velocity limits can be encoded into the QP (Eq. \ref{eq:velocity_cbf}) via an additional \(2n_q\) constraints:
\begin{equation}
    \dot{\mathbf{q}}_{\text{min}} \leq \dot{\mathbf{q}} \leq \dot{\mathbf{q}}_{\text{max}}
    \label{eq:joint_vel_constraint}
\end{equation}

\textit{Remark}: Adding input constraints to this QP does not guarantee forward invariance of the safe set, but it often still works in practice without introducing more complexity. For a more thorough handling, see input-constrained CBFs \cite{AgrawalInputConstrainedCBFs}.

The optimizer of the QP (Eq. \ref{eq:velocity_cbf}), \(\dot{\mathbf{q}}^*\), can then be safely sent to the robot.

\subsection{Dynamic-OSCBF}

\begin{figure*}
    \centering
    \smallskip
    \includegraphics[width=0.75\linewidth]{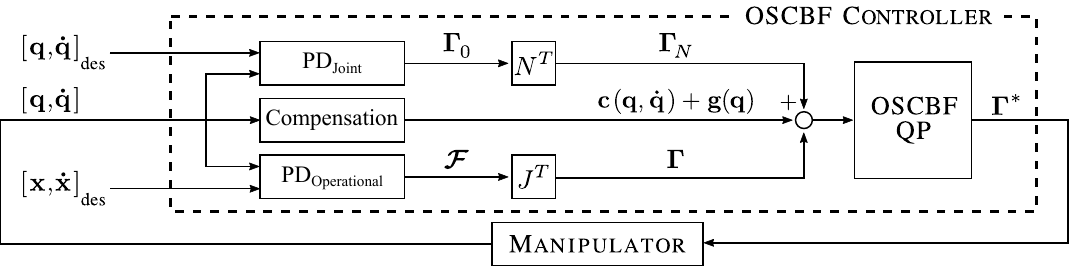}
    \caption{\textbf{Block diagram}: OSCBF for torque-controlled manipulators}
    \label{fig:block_diagram}
    \vspace{-2mm}
\end{figure*}

For torque-controlled manipulators, computing the safe control input \(\boldsymbol{\Gamma}\) requires three steps: (1) Determine the nominal generalized forces to reduce error in the task hierarchy; (2) Map the nominal generalized forces to an (unsafe) joint torque command; (3) Pass the nominal joint torque command through the OSCBF QP to yield the safe torque command \(\boldsymbol{\Gamma}^*\). A block diagram of this process can be found in Fig. \ref{fig:block_diagram}.

We begin by computing the nominal generalized forces for both the operational and joint space tasks,
\begin{equation}
    \boldsymbol{\mathcal{F}} =  \Lambda (\mathbf{q})\dot{\boldsymbol{\nu}}
\end{equation}
\begin{equation}
    \boldsymbol{\Gamma}_0 = M(\mathbf{q})\ddot{\mathbf{q}}
\end{equation}
where \(\boldsymbol{\mathcal{F}}\) is an operational space wrench command \(\in \mathbb{R}^6\) and \(\boldsymbol{\Gamma}_{0}\) is a joint-space torque command \(\in \mathbb{R}^{n_q}\).

The nominal operational and joint space accelerations, \(\dot{\boldsymbol{\nu}}\) and \(\ddot{\mathbf{q}}\), are computed via proportional-derivative (PD) control of the task error dynamics:
\begin{equation}
    \dot{\boldsymbol{\nu}}
    =
    \dot{\boldsymbol{\nu}}_{\text{des}} 
    - K_{po} 
    \begin{bmatrix} 
        \mathbf{x}_p - \mathbf{x}_{p,\text{des}}\\  
        \boldsymbol{\delta\phi} 
    \end{bmatrix} 
    - K_{do} (\boldsymbol{\nu} - \boldsymbol{\nu}_{\text{des}})
\end{equation}
\begin{equation}
    \ddot{\mathbf{q}}
    =
    \ddot{\mathbf{q}}_{\text{des}} - K_{pj}(\mathbf{q} - \mathbf{q}_{\text{des}}) - K_{dj}(\dot{\mathbf{q}} - \dot{\mathbf{q}}_{\text{des}})
\end{equation}
where \(K_{po}, K_{do},K_{pj}, K_{dj}\) are the PD gains for the operational and joint spaces, respectively.

Given the task hierarchy, the joint space task is then projected into the null space of the operational space task:
\begin{equation}
    \boldsymbol{\Gamma}_N = N^T(\mathbf{q})\boldsymbol{\Gamma}_0
\end{equation}

Combining these together, the nominal (unsafe) torque command, with gravity and centrifugal/Coriolis compensation is
\begin{equation}
    \boldsymbol{\Gamma}_{\text{nom}} = J^T(\mathbf{q}) \boldsymbol{\mathcal{F}} + \boldsymbol{\Gamma}_N + \mathbf{c}(\mathbf{q}, \dot{\mathbf{q}}) + \mathbf{g}(\mathbf{q})
\end{equation}

To construct the CBF, we first define the control-affine joint-space dynamics (\(\mathbf{z} = \left[ \mathbf{q}, \dot{\mathbf{q}} \right] \in \mathbb{R}^n\)), with respect to the joint torques (\(\mathbf{u} =\boldsymbol{\Gamma} \in \mathbb{R}^m\)):
\begin{equation}
    \dot{\mathbf{z}} = 
    \begin{bmatrix}
        \dot{\mathbf{q}} \\
        -M^{-1}(\mathbf{q})(\mathbf{c}(\mathbf{q}, \dot{\mathbf{q}}) + \mathbf{g}(\mathbf{q}))\\
    \end{bmatrix}
     + 
    \begin{bmatrix} 
        \mathbf{0} \\
        M^{-1}(\mathbf{q})  \\
    \end{bmatrix}
    \mathbf{u}
\end{equation}

With these dynamics, and a CBF \(h\), we can then construct the OSCBF QP,
\begin{equation}
    \begin{aligned}
        \underset{\boldsymbol{\Gamma}}{\text{minimize}} \quad & \| W_j (\ddot{\mathbf{q}}_{N} - \ddot{\mathbf{q}}_{N\text{nom}}) \|_2^2 +  \| W_o (\dot{\boldsymbol{\nu}} - \dot{\boldsymbol{\nu}}_{\text{nom}}) \|_2^2 \\ %
        \text{subject to} \quad & L_f h(\mathbf{z}) + L_g h(\mathbf{z}) \mathbf{u} \geq -\alpha\left( h(\mathbf{z}) \right) \\ %
    \end{aligned}
    \label{eq:torque_cbf}
\end{equation}

As with the OSCBF QP for kinematic control, the CBF constraint is paired with a task-consistent objective, though here, we minimize the deviations from the nominal joint and operational space \textit{accelerations}. \(W_j\) and \(W_o\) are the joint/operational space weighting matrices, from before. Note: if the CBF is of relative degree 2, we instead apply a HOCBF constraint as in Eq. \ref{eq:rd2_cbf_constraint}.

Equivalently, the objective function of Eq. \ref{eq:torque_cbf} can be written as
\begin{multline}
    \| W_j M^{-1}(\mathbf{q}) N^T(\mathbf{q})(\boldsymbol{\Gamma} - \boldsymbol{\Gamma}_{\text{nom}}) \|_2^2 \\ +  \| W_o J(\mathbf{q})M^{-1}(\mathbf{q})(\boldsymbol{\Gamma} - \boldsymbol{\Gamma}_{\text{nom}}) \|_2^2
\end{multline}
which we can then use to construct a QP with an objective of form \(\frac{1}{2}\mathbf{x}^T P_{QP} \mathbf{x} + \mathbf{q}_{QP}^T\mathbf{x}\), where
\begin{equation}
    P_{QP} = N M^{-T} W_j^T W_j M^{-1} N^T + M^{-T} J^T W_o^T W_o J M^{-1}
\end{equation}
\begin{equation}
    \mathbf{q}_{QP}^T = -\boldsymbol{\Gamma}_{\text{nom}}^T P_{QP}
\end{equation}

Similarly to Eq. \ref{eq:velocity_hierarchy}, this objective can also generalize to a larger hierarchy, by considering the effect of the safety filter on each task \(i\):
\begin{equation}
    \sum_{i=1}^{n_t} \Vert W_i  (\mathbf{a}_i(\mathbf{q}, \dot{\mathbf{q}}, \boldsymbol{\Gamma}) - \mathbf{a}_{i, \text{nom}}(\mathbf{q}, \dot{\mathbf{q}}, \boldsymbol{\Gamma}_{\text{nom}}))\Vert_2^{2}
\end{equation}
where \(\mathbf{a}_i\) is an acceleration, in operational or joint space.

Joint torque limits can be encoded into the QP (Eqn. \ref{eq:torque_cbf}) via an additional \(2n_q\) constraints: 
\begin{equation}
    \boldsymbol{\Gamma}_{\text{min}} \leq \boldsymbol{\Gamma} \leq \boldsymbol{\Gamma}_{\text{max}}
\end{equation}

Operational space contact wrench limits (\(||\boldsymbol{\mathcal{F}}_c||_2 \leq \boldsymbol{\mathcal{F}}_{c,\text{max}}\)) can be encoded via the following linear inequality approximation: 
\begin{equation}
    \boldsymbol{\mathcal{F}}_{c,\text{min}} \leq \bar{J}^T(\mathbf{q})\left(\boldsymbol{\Gamma} - \mathbf{c}(\mathbf{q}, \dot{\mathbf{q}}) - \mathbf{g}(\mathbf{q})  \right) \leq \boldsymbol{\mathcal{F}}_{c,\text{max}}
\end{equation}

As noted for kinematic control (Eq. \ref{eq:joint_vel_constraint}), input constraints do not guarantee safety if they are not designed into the CBF itself, but can still work well in practice. For high-relative-degree input-constrained CBFs, see \cite{BreedenHighRelativeDegInputConstraints}.

The optimizer of the QP (Eq. \ref{eq:torque_cbf}), \(\boldsymbol{\Gamma}^*\), can then be safely sent to the robot.

\section{\label{sec:results}Experiments and Results}

\subsection{Implementation}

\begin{table*}
    \centering
    \smallskip
    \caption{OSCBF Control Frequencies (kHz): 7-DOF Franka Emika Panda}
    \label{tab:speed}
    \begin{tabular}{@{\extracolsep{4pt}}lccccccc@{}}
        \toprule
        \multicolumn{1}{c}{Experiment}         & CBF Constraints & \multicolumn{2}{c}{Velocity Control} & \multicolumn{2}{c}{Torque Control}  \\
        \cmidrule{3-4} \cmidrule{5-6} \cmidrule{7-8}
                                           &      & Mean   & Lower bound (5th percentile)   & Mean  & Lower bound (5th percentile)  \\\midrule
        Singularity avoidance              & 1    & 9.55   & 5.15                    & 6.14  & 3.89 \\
        End-effector safe-set containment  & 6    & 9.49   & 6.19                    & 7.17  & 4.12 \\
        Joint limit avoidance              & 14   & 9.99   & 6.15                    & 7.72  & 4.45 \\
        Whole-body collision avoidance     & 21   & 7.48   & 4.40                    & 5.71  & 3.06 \\
        Whole-body safe-set containment    & 126  & 4.94   & 3.33                    & 3.89  & 2.59 \\
        (All of the above)                 & 168  & 3.24   & 2.35                    & 2.94  & 2.25 \\
        \bottomrule
    \end{tabular}
    \vspace{-2mm}
\end{table*}

We present CBFpy \cite{Morton_CBFpy_2024}: a high-performance software package for constructing and solving CBFs. We use Jax \cite{jax2018github} for automatic differentiation of the barrier functions, just-in-time (JIT) compilation of the Python implementation into high-performance XLA code \cite{XLA}, and efficient Jacobian-vector-products for the Lie derivatives. The QP is solved via a primal-dual interior point method, also in Jax \cite{TracyQPax}. 

The Python interface is sufficiently fast for kilohertz control rates (see Table \ref{tab:speed}), even when the full manipulator dynamics are evaluated, and with over 100 CBF constraints. 
If C++ code is required, the OSCBF can be directly converted to a high-level optimized XLA representation and called via the XLA C++ interface. 
Compute times are from an Intel i7-1360p (5 GHz) NUC computer with 64 GB RAM, and Jax v.0.4.30. %
JIT compilation times are typically 2-5 seconds.

\subsection{Multiple Safety Constraints}

Manipulator control often requires handling multiple safety constraints simultaneously. For our experiments, we consider five common types of constraints: (1) singularity avoidance, (2) end-effector safe-set containment, (3) joint limit avoidance, (4) whole-body collision avoidance, and (5) whole-body safe-set containment. 

To validate that safety is enforced for all constraints, we command an (unsafe) teleoperated end-effector trajectory that approaches the limit of safety for each constraint (Fig. \ref{fig:combined}). Each constraint is driven towards the boundary of safety (\(h(\mathbf{z}) = 0\)), yet none enter the unsafe region (\(h(\mathbf{z}) < 0\)), and no conditions restrict the robot motion prematurely, indicating excellent tracking performance, without over-conservatism in the face of safety.  Additionally, even with 168 CBF constraints enforced concurrently, the OSCBF controller is \textit{fast}: retaining over 1000 Hz and real-time control rates for velocity and torque control (Table \ref{tab:speed}).

Below, we indicate the CBF construction for each constraint. Note: the following CBFs are of relative degree 1 for velocity control, and relative degree 2 for torque control. For torque control, we apply Eq. \ref{eq:rd2_cbf_constraint} to resolve the relative degree.

\subsubsection{Singularity avoidance}
As in \cite{SingularityCBF}, we define the barrier function based on the manipulability index \(\mu(\mathbf{q})\) \cite{Manipulability} and a tolerance \(\epsilon\),
\begin{equation}
    h(\mathbf{z}) = \mu(\mathbf{q}) - \epsilon
\end{equation}
where \(\mu(\mathbf{q})\) is equal to the product of the singular values \(\sigma\) of the end-effector Jacobian, 
\begin{equation}
    \mu(\mathbf{q}) = \prod_{i=1}^{6} \sigma_i
\end{equation}

\subsubsection{Joint position limit avoidance}
We construct the joint limit CBF as simply the joint-space distance to the minimum and maximum values,
\begin{equation}
    h(\mathbf{z}) =
    \begin{bmatrix}
        \mathbf{q} - \mathbf{q}_{\text{min}}\\
        \mathbf{q}_{\text{max}} - \mathbf{q} \\
    \end{bmatrix}
\end{equation}

\subsubsection{Task position limit avoidance}
We assume an axis-aligned bounding box representation of the end-effector safe set, with a barrier function
\begin{equation}
    h(\mathbf{z}) =
    \begin{bmatrix}
        \mathbf{x}_{p,EE}(\mathbf{q}) - \mathbf{x}_{p,\text{min}}\\
        \mathbf{x}_{p,\text{max}} - \mathbf{x}_{p,EE}(\mathbf{q}) \\
    \end{bmatrix}
\end{equation}
where \(\mathbf{x}_{p,EE}(\mathbf{q})\) represents the position of the end-effector, after propagation through the forward kinematics. 

\textit{Remark}: Barrier functions can be constructed for a variety of keep-in (or keep-out) zones, not just axis-aligned boxes. Ellipsoids, halfspaces, polytopes, capsules, and more are all possible. For further discussion, refer to \cite{tracy2023differentiable,DaiDiffOptCBF}.

\subsubsection{Whole-body collision avoidance}
We assume a sphere-decomposition collision model of the robot and the environment, with barrier functions
\begin{equation}
    h_{ij}(\mathbf{z}) = ||\mathbf{x}_{p,j,\text{obs}} - \mathbf{x}_{p,i}(\mathbf{q})||_2 - r_{j,\text{obs}} - r_{i}
    \label{eq:collision_cbf}
\end{equation}
for all \(i\) positions \(\mathbf{x}_{p,i}(\mathbf{q})\) and radii \(r_{i}\) in the collision model of the robot, after propagation through the forward kinematics, and all \(j\) positions/radii of the environmental collision model, \(\mathbf{x}_{p,j,\text{obs}}\) and \(r_{j,\text{obs}}\). Our simple collision model for the Franka Panda consists of 21 spheres, and we emphasize that this will still work with a more detailed collision model. 

\subsubsection{Whole-body set containment}
We assume an axis-aligned bounding box representation of the whole-body safe set, and again a sphere-based collision model of the robot.
\begin{equation}
    h_{i}(\mathbf{z}) =
    \begin{bmatrix}
        \mathbf{x}_{p,i} - \mathbf{x}_{p,\text{min}} - r_i \\
        \mathbf{x}_{p,\text{max}} - \mathbf{x}_{p,i} - r_i
    \end{bmatrix}
    \label{eq:whole_body_set_cbf}
\end{equation}

\subsection{Highly-Constrained Environments}

\begin{figure}
    \centering
    \smallskip
    \includegraphics[width=\linewidth]{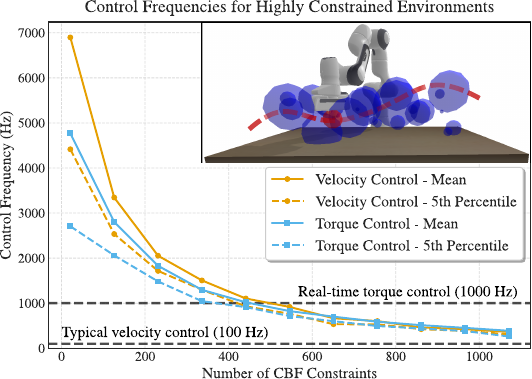}
    \caption{\textbf{Scaling up collision avoidance}. Even in highly-cluttered scenes, our OSCBF controller maintains safety, task-tracking performance, and real-time control rates. Consider a tabletop environment (inset) with many randomly-generated collision bodies, shown in blue. OSCBF scales to \textit{over 400 CBF constraints while retaining real-time control rates} (1000 Hz) for torque control, and well over 1000 constraints while retaining good control rates (100 Hz) for velocity control. We indicate both the mean frequency and the minimum frequency, assuming an allowable 5\% packet drop.}
    \label{fig:scaling_collisions}
    \vspace{-2mm}
\end{figure}

Operating in an unstructured environment such as the household often implies a lot of clutter: to name a few, reaching into narrow cabinets filled with items, moving dishes in and out of the sink, or cleaning up laundry \cite{khazatsky2024droid}. %
To validate the performance of the controller in a similar setting, we consider a cluttered tabletop environment, where we enforce collision avoidance with both the table and a variable number of bodies (0 to 50), represented as a randomly generated set of spherical regions. The table is modeled with a halfspace CBF (as in Eq. \ref{eq:whole_body_set_cbf}, but only along the \(z\) direction), and each additional CBF constraint (Eq. \ref{eq:collision_cbf}) pairs one collision sphere on the robot with one collision sphere in the environment. Given the 21 spherical bodies in our simplified collision model of the Franka Panda, this implies that we have more than 1000 CBF constraints when we enforce whole-body collision avoidance with 50 environmental bodies. 

As shown in Fig. \ref{fig:scaling_collisions}, we retain real-time control rates (1000 Hz) for over 400 CBF constraints, and maintain good control rates for velocity control (well over 100 Hz) even with over 1000 CBF constraints in the QP. We track an end-effector trajectory that safely moves through even the most tightly-occluded sections of the environment with no collisions -- a task that would be infeasible for APFs due to the interference between repulsive potentials, and infeasible for MPC due to the high number of nonconvex constraints. Notably, we can ensure safety even with dynamic motions through the cluttered scene, given the real-time performance of torque control, with centrifugal and Coriolis compensation. 

\textit{Remark}: We can further increase computational efficiency by considering only the closest subset of collision pairs: this simple heuristic can dramatically raise the number of collision bodies in the environment for a given compute frequency. However, results show that even without these heuristics, we can enforce whole-body collision avoidance in high levels of clutter.

\subsection{Dynamic Safety}

\begin{figure}
    \centering
    \smallskip
    \includegraphics[width=\linewidth]{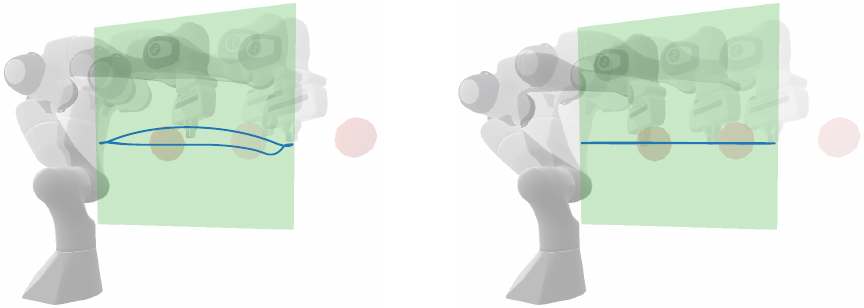}
    \caption{\textbf{Dynamic task consistency under input constraints}. Under high-speed unsafe motions, accounting for the full dynamics of the robot and torque input constraints is necessary for good task performance. Consider a periodic, straight-line end-effector trajectory that commands a rapid motion of the end-effector tip into the unsafe set. With constraints on both the maximum joint velocities and torque, both a velocity-controlled robot (left) and a torque-controlled robot (right) maintain safety. However, the velocity CBF causes a degradation of tracking performance, due to instantaneously infeasible velocity commands, given the configuration of the robot and its torque limits.}
    \label{fig:dynamic_safety}
    \vspace{-2mm}
\end{figure}

Manipulators are often capable of high-speed dynamic motions, but guaranteeing safety during these motions can be difficult, particularly in an online setting without pre-planned trajectories. 
Previous works assumed that a low-level tracking controller can guarantee that a safe velocity command can always be met, but we emphasize that this is \textit{not} the case for dynamic motions, with torque limits. As shown in Fig. \ref{fig:dynamic_safety}, making this assumption can sometimes maintain safety (here, the end-effector trajectory does remain within the safe set), but the desired safe velocity reference cannot always be met, given the torque limits. This leads to a degradation in task tracking performance, as indicated by the departure from the desired straight-line motion. However, if the full second-order dynamics of the manipulator are considered, with explicit constraints on the torque input, we can maintain \textit{both} tracking performance and safety.

\subsection{Hardware Experiments}

\begin{figure}
    \centering
    \smallskip
    \includegraphics[width=\linewidth]{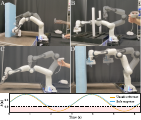}
    \caption{\textbf{Hardware validation}. We evaluate OSCBF's performance across four scenarios on the Franka Emika Panda hardware. In three teleoperated experiments, we demonstrate (A) whole-body workspace containment, (B) whole-body collision avoidance, and (C) singularity avoidance. In (D), we command a fast sinusoidal end-effector trajectory which, without OSCBF, would lead directly to a collision. The plot visualizes the minimum CBF constraint evolution in this experiment, indicating smooth, safe, and non-overly-conservative behavior from the OSCBF controller. 
    }
    \label{fig:hardware}
    \vspace{-2mm}
\end{figure}

To validate OSCBF's performance on hardware, we set up a 1 kHz real-time torque control interface for safe teleoperation and operational space control of the Franka Panda. Across four experiments designed to replicate the simulation results, OSCBF demonstrates remarkably smooth, low-latency tracking of the end-effector commands, safely approaching the boundary of safety within just 6 mm (Fig. \ref{fig:hardware} A/B/D) or quickly moving in and out of near-singular configurations (\(\mu = 1\mathrm{e}{-2}\)) (Fig. \ref{fig:hardware} C) while operating at the dynamic limits of the Franka hardware. Fig. \ref{fig:hardware} B also demonstrates whole-body collision avoidance while maintaining good tracking of the operational space task, by filtering the secondary null space posture task when constrained by the environment. Notably, we observe effectively zero mismatch between the hardware data and simulation, save for a miniscule amount of latency (50 ms) from the teleoperation interface. Videos of the experiments can be found on our project webpage.

\subsection{Discussion}

\subsubsection{Parameter tuning}
For a good balance between performance and conservatism in safety constraints, the controller gains \(K_{po}, K_{do}, K_{pj}, K_{dj}\), CBF class \(\mathcal{K}_\infty\) functions \(\alpha, \alpha_2\) (for each type of CBF), and objective weights \(W_j, W_o\) can be tuned. In general, the designer should (1) tune the controller gains to achieve good unconstrained performance, (2) enforce constraints and raise \(\alpha, \alpha_2\) starting from 1 until velocity/torque limits are nearly reached under dynamic motions, and (3) adjust \(W_j, W_o\) starting from \(I\) to relax or strengthen deviations in end-effector position, rotation, or joint motions. \textit{Remark}: OSCBF works even with minimal tuning. For all experiments, \(\alpha = \alpha_2 = 10\) (for all CBFs) and \(W_j = I_{n_q}\), \(W_o = I_{6}\) performed well.

\section{Conclusion}

In this work, we have introduced Operational Space Control Barrier Functions as a real-time control method for enforcing safety in hierarchical tasks, in both the operational and joint spaces. By defining the CBF objective in a task-consistent manner, we retain a high level of performance even when the control input is constrained by the safety filter, avoiding both over-conservatism and unnecessary motions. We then scale up the number of CBFs, demonstrating kilohertz control rates even with hundreds of constraints in the QP. This allows for safety even under dynamic motions, or in extremely cluttered environments.

Future work will focus on applications on high-DOF mobile manipulators or bimanual systems, and integration with real-time perception. Moving forward, we aim to apply this controller as a core part of imitation-learning-based manipulation policy training and deployment: collecting data safely via teleoperation, and deploying the learned policies with OSCBF maintaining whole-body safety of the robot.

\bibliographystyle{IEEEtran}
\bibliography{references}

\section{Appendix}

\subsection{Additional Safety Constraints}

\subsubsection{Velocity Limits}

For a torque-controlled manipulator, joint velocity limits can be enforced as a relative degree 1 CBF, with the following form:
\begin{equation}
    h(\mathbf{z}) =
    \begin{bmatrix}
        \dot{\mathbf{q}} - \dot{\mathbf{q}}_{\text{min}}\\
        \dot{\mathbf{q}}_{\text{max}} - \dot{\mathbf{q}} \\
    \end{bmatrix}
\end{equation}

Similarly, for an operational space velocity or twist limit, we can define
\begin{equation}
    h(\mathbf{z}) =
    \begin{bmatrix}
        \dot{\boldsymbol{\nu}}(\mathbf{q}, \dot{\mathbf{q}}) - \dot{\boldsymbol{\nu}}_{\text{min}}\\
        \dot{\boldsymbol{\nu}}_{\text{max}} - \dot{\boldsymbol{\nu}}(\mathbf{q}, \dot{\mathbf{q}}) \\
    \end{bmatrix}
\end{equation}

where \(\dot{\boldsymbol{\nu}}(\mathbf{q}, \dot{\mathbf{q}})\) is defined by the manipulator kinematics, Eq. \ref{eq:kinematic_model}.

For a velocity-controlled manipulator, these must instead be included as a constraint in the QP (see Table \ref{tab:constraint_or_cbf}).

\subsubsection{Self Collision}

As with the whole-body collision avoidance and set containment CBFs (Eqs. \ref{eq:collision_cbf}, \ref{eq:whole_body_set_cbf}), we assume a sphere-based collision model of the robot. 

Given a set of self-collision pairs of sphere indices \(p_i = (j, k)\), we form the barrier function
\begin{equation}
    h_{i}(\mathbf{z}) = ||\mathbf{x}_{p,j}(\mathbf{q}) - \mathbf{x}_{p,k}(\mathbf{q})||_2 - r_{j} - r_{k}
\end{equation}

for all \(i\) pairs. As before, \(\mathbf{x}_p(\mathbf{q})\) represents the position of a sphere after propagation through the forward kinematics, and \(r\) is the corresponding radius of the sphere. 

\textit{Remark:} The self-collision model and the whole-body collision avoidance models do not need to have the same geometry. The self-collision model can be a much simpler representation to account for only the possible collision modes, given the robot's kinematics.

\subsubsection{Dynamic Obstacles}

For a dynamic obstacle, we can form the following barrier function, again assuming spherized representations of the obstacle (\(o\)) and the robot's \(i^{th}\) collision body:
\begin{equation}
    h_{i}(\mathbf{z}) = ||\mathbf{x}_{p,i}(\mathbf{q}) - \mathbf{x}_{p,o}||_2 - \gamma ||\mathbf{v}_{i,\text{rel}}||_2 - r_{i} - r_{o} 
\end{equation}

Here, \(||\mathbf{v}_{i,\text{rel}}||_2\) represents the norm of the relative velocity between the obstacle and the robot's \(i^{th}\) collision body, and \(\gamma\) is a scaling factor. This has the effect of inflating the obstacle by an amount proportional to the relative velocity.

\textit{Remark:} This is just one of several possible strategies for handling dynamic obstacles that was found to perform well in practice. Here, the scaling factor \(\gamma\) dictates the conservatism of the robot around moving obstacles, similar to how \(\alpha\) can be used to tune conservatism. For the previously-mentioned choice of \(\alpha = \alpha_2 = 10\), \(\gamma = 0.25\) performed well.

\end{document}